\documentclass{article}

\usepackage[preprint]{neurips_2026}

\usepackage[utf8]{inputenc} 
\usepackage[T1]{fontenc}    

\usepackage{hyperref}       
\usepackage{url}            
\usepackage{booktabs}       
\usepackage{amsfonts}       
\usepackage{nicefrac}       
\usepackage{microtype}      
\usepackage{xcolor}         
\usepackage{booktabs}       
\usepackage{graphicx}
\usepackage{amsmath}
\usepackage{amssymb}
\usepackage{array}
\usepackage{multirow}
\usepackage{color}
\usepackage{colortbl}
\usepackage{framed}
\usepackage{bm}
\usepackage{xspace}
\usepackage{enumitem}
\usepackage{xparse}
\usepackage{algorithm}
\usepackage{listings}
\usepackage{url}
\usepackage{mathtools}
\usepackage{pifont}
\usepackage{mathtools}
\usepackage{lipsum}
\usepackage{adjustbox}
\usepackage{bbm}

\usepackage{multirow}
\usepackage{graphicx} 
\usepackage[table]{xcolor}
\usepackage{algpseudocode}
\usepackage{soul} 
\usepackage{mathtools}
\usepackage{xcolor}
\usepackage{mathtools}
\usepackage{pifont}
\usepackage{mathtools}
\usepackage{lipsum}
\usepackage{adjustbox}
\usepackage{bbm}
\usepackage{makecell}
\usepackage[x11names]{xcolor}

\definecolor{Gray}{gray}{0.2}
\definecolor{lightgray}{gray}{0.92}
\definecolor{blond}{rgb}{0.98, 0.94, 0.75}
\definecolor{TitleColor}{gray}{0.95}
\definecolor{LightCyan}{rgb}{0.88,0.95,1}
\definecolor{OurColor}{rgb}{0.82, 0.88, 0.97}

\definecolor{blond}{rgb}{0.98, 0.94, 0.75}
\def \ie {\emph{i.e.}}
\def \eg {\emph{e.g.}}
\def \etal {\emph{et al.}}

\newcommand{\tit}[1]{\noindent\textbf{#1.}}

\definecolor{MyGreen}{RGB}{34,139,34}

\definecolor{Gray}{gray}{0.93}
\definecolor{tablegray}{RGB}{122, 122, 122}

\title{Do Models Share Safety Representations? \\Cross-Model Steering for Safe Visual Generation}

\author{%
  \textbf{Tobia Poppi}\textsuperscript{1,2} \quad
  \textbf{Silvia Cappelletti}\textsuperscript{1} \quad
  \textbf{Sara Sarto}\textsuperscript{1} \quad
  \textbf{Florian Schiffers}\textsuperscript{3} \\
  \textbf{Garin Kessler}\textsuperscript{3} \quad
  \textbf{Marcella Cornia}\textsuperscript{1} \quad
  \textbf{Lorenzo Baraldi}\textsuperscript{1} \quad
  \textbf{Rita Cucchiara}\textsuperscript{1} \\
  \\
    \textsuperscript{1}University of Modena and Reggio Emilia \quad \textsuperscript{2}University of Pisa \quad \textsuperscript{3}Amazon Prime Video
  \\
  \\
  {\tt\small \href{https://aimagelab.github.io/cross-model-safety-representations/}{aimagelab.github.io/cross-model-safety-representations}}
  \vspace{-0.5cm}
}

\begin{document}

\maketitle

\begin{abstract}
Recent progress in generative modeling has made safety control a central challenge, yet existing approaches remain largely model-specific, requiring retraining or tailored interventions for each new architecture. In this work, we ask whether safety can be represented as a \emph{portable latent direction}, learned once and reused across heterogeneous generators. We introduce the first framework for \emph{cross-model safety steering}, in which a safety direction is estimated in a source LLM from paired safe-unsafe prompts, transported to a target generator through a lightweight alignment fitted on benign data alone, and applied at inference time. Crucially, our pipeline never accesses unsafe data on the target side, isolating whether safety can be transferred through shared representation geometry. Beyond a single global direction, we also identify a multi-vector extension that captures category-specific safety behaviors, enabling more selective control. We evaluate our approach in text-to-image and text-to-video generation across diverse source-target model pairs.
Across models, transferred safety directions achieve ASR reduction and CLIP-Score/FID trade-offs comparable to directions learned natively on the target model using unsafe data, while requiring no target-side unsafe data. This indicates that safety improvements do not come at the expense of generation quality. Our results point to a modular view of safety: safety-relevant behavior is not purely model-local, but can be controlled through latent directions that persist across models. This suggests a new path toward lightweight, reusable safety mechanisms that do not require target-side unsafe data.

\medskip
{\color{Firebrick3} \noindent\textit{\textbf{Warning:} This paper contains examples of harmful and explicit content, including sexual and violent material, which some readers may find disturbing or offensive.}\vspace{-0.4cm}}
\end{abstract}

\begin{figure}[H]
    \centering
    \includegraphics[width=0.98\linewidth]{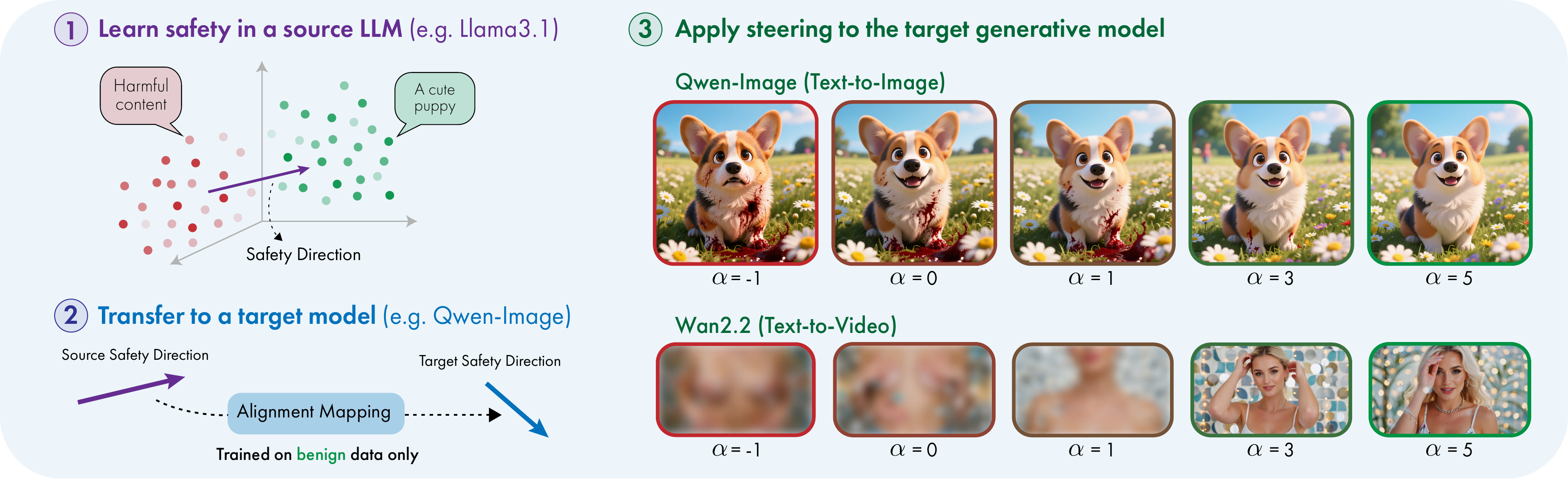}
    \vspace{-0.1cm}
    \caption{Safety as a transferable direction. A safety vector learned in a source model is aligned to a target model using only benign, safe data and applied at inference time to steer generation. The right panel illustrates the resulting safety behavior on Qwen-Image and Wan2.2 as the steering strength $\alpha$ varies, where unsafe content is progressively suppressed while scene composition is preserved.}
    \label{fig:teaser}
    \vspace{-0.3cm}
\end{figure}

\section{Introduction}
\label{sec:introduction}

Safety has become a central challenge in modern generative modeling. LLMs~\cite{zhao2023survey}, text-to-image systems~\cite{esser2024scaling,flux2024,rombach2022high}, and video generators~\cite{wan2025wan} are trained on massive, only partially curated corpora, and inevitably internalize unsafe or policy-sensitive concepts, including sexual content, hate, violence, illegal activity, and copyrighted or trademarked entities~\cite{abid2021persistent,birhane2021multimodal,bommasani2021opportunities,weidinger2022taxonomy}. As these systems are increasingly deployed, there is a growing need for safety mechanisms that are both effective and lightweight.

A growing body of work suggests that safety is not only an output-level property, but is also reflected in the geometry of internal representations~\cite{bello2025linear,huang2025cross,stolfo2025improving}. In large language models, representation engineering has shown that harmful or refusal-related behavior can often be associated with identifiable directions in activation space, and that intervening on these internal states can alter model behavior~\cite{arditi2024refusal,zou2024improving}. Similar observations in diffusion-based image and video models indicate that safety can be improved through latent-space interventions, rather than only through prompt filtering or retraining~\cite{facchiano2026videounlearning,poppi2024safe,schramowski2023safe}. Collectively, these results suggest that safety-relevant behavior may be encoded as structured directions in representation space.

At the same time, recent work on representation similarity indicates that learned latent spaces across models are often not independent, but exhibit nontrivial geometric alignment and can sometimes be related by simple mappings~\cite{jha2025universal}. This perspective is consistent with the \emph{Platonic Representation Hypothesis}, which posits that sufficiently capable models converge toward shared internal representations of high-level concepts~\cite{huh2024platonic}.

In this paper, we bring these two lines of work together and ask: \emph{are safety representations transferable across models?} Specifically, we investigate whether a safety direction learned in a source model remains meaningful when mapped into the latent space of a different target model, even when source and target differ in architecture, training data, pretraining paradigm, or tokenizer, and without access to unsafe data in the target. This setting reflects a common practical constraint, where collecting or using unsafe target data is infeasible or undesirable, yet safety adaptation is still required.

To address this question, we propose a simple and general framework for \emph{cross-model safety steering}. We first estimate a safety vector in a source model from paired safe-unsafe prompts, given by the direction from harmful to safe representations. We then learn a lightweight alignment between source and target representation spaces using only safe anchor prompts, and transport the safety direction into the target model. Finally, we apply the transferred direction at inference time to steer generation. Beyond a single global direction, we extend this formulation to a multi-vector setting, where category-specific safety directions are estimated and transferred independently, enabling more selective control over different types of unsafe behavior.
Fig.~\ref{fig:teaser} provides an overview of this cross-model transfer process and illustrates its effect as the steering strength is varied.
Our formulation isolates a key question: does safety transfer require model-specific supervision, or can it emerge from shared representation geometry? By restricting alignment to benign, safe data and using simple mappings, our approach directly tests whether safety-relevant structure is portable rather than model-specific.

Empirically, we evaluate this framework in text-to-image generation across diverse source (\ie, Llama3.1~\cite{grattafiori2024llama}, Mistral~\cite{jiang2023mistral7b}, and Qwen3.5~\cite{qwen35blog}) and target models (\ie, Flux1-Schnell, Flux1-Dev~\cite{flux2024}, Qwen-Image~\cite{wu2025qwen}, Z-Image-Turbo~\cite{cai2025z}), and extend the analysis to text-to-video generation (using Wan2.2~\cite{wan2025wan} as target model). We measure both safety (via attack success rate~\cite{bedapudi2019nudenet,schramowski2022can}) and utility (via CLIP similarity~\cite{radford2021learning} and FID~\cite{heusel2017gans}), enabling a fine-grained characterization of the safety-utility trade-off. Our results show that safety directions learned in one model can be transferred across heterogeneous generative models using only safe alignment data, substantially reducing unsafe generations while preserving performance on safe content.

These findings suggest a modular view of safety: safety directions and alignment statistics can be learned once and reused across models, without exposing sensitive data. Beyond practical implications, our results provide evidence that safety-relevant concepts are not purely model-local, but are grounded in shared geometric structure across generative systems.
\section{Related Work}
\label{sec:related}
We review prior work along four main directions: \textit{(i)} safety alignment in generative models, \textit{(ii)} activation steering, \textit{(iii)} representation alignment, and \textit{(iv)} cross-model transfer of interventions.

\tit{Safety Alignment in Generative Models}
In text-to-image generation, prior work includes model editing methods that suppress unsafe content or steer models toward safer outputs~\cite{fan2023salun,gandikota2023erasing,gandikota2024unified,huang2024receler,liu2025safetydpo,lu2024mace,lyu2024one}, guidance-based approaches applied at sampling time~\cite{schramowski2023safe}, and pre-decoding methods that sanitize or align representations before generation~\cite{ahn2025des,liu2024latent,poppi2024safe}.  More recently, activation-level control has emerged as a lightweight inference-time alternative. AcT~\cite{rodriguezcontrolling} learns an affine map that transports activations from a source distribution to a target distribution, enabling control over toxicity, concept induction, and truthfulness across both language and diffusion models. CAT~\cite{chrabkaszcz2026conditioned} extends this approach to text-to-image safety by learning conditioned nonlinear transport maps that act primarily in unsafe regions of the activation space.

\tit{Activation Steering}
A related line of work investigates whether high-level behaviors correspond to linear directions in activation space. Early studies show that targeted activation perturbations can reliably steer model behavior~\cite{subramani2022extracting,turner2023steering}. Representation engineering~\cite{zou2023representation} and theoretical analyses~\cite{park2024linear} further support the hypothesis that semantic features exhibit linear structure in representation space. Building on this view, CAA~\cite{rimsky2024steering} derives steering vectors from contrastive activation differences to modulate behaviors such as sycophancy and refusal, while Arditi~\etal~\cite{arditi2024refusal} identify a dominant refusal direction that can be added or removed to control safety behavior. Similar observations extend to video generation, where safety-relevant directions can be extracted from paired data, but are primarily used for within-model edits rather than transferable interventions~\cite{facchiano2026videounlearning}. Collectively, these results suggest that safety behaviors can often be captured as latent directions.

\tit{Representation Alignment}
Our method also relies on the premise that simple learned mappings can relate internal representations from different models. Early model stitching work showed that independently trained networks can be connected through lightweight transformations~\cite{bansal2021revisiting,lenc2015understanding}.
The \textit{Platonic Representation Hypothesis}~\cite{huh2024platonic} further suggests that different models may converge toward shared representational structure.
Empirically, Jha~\etal~\cite{jha2025universal} demonstrate that embeddings can be translated across models without paired data via shared latent structure, while Chen~\etal~\cite{chen2025transferring} show that affine maps can transfer sparse autoencoders, probes, and steering vectors across language models, supporting the portability of linear features.

\tit{Cross-Model Intervention Transfer}
More recent work has started to explicitly study the transfer of interventions across models. Stolfo~\etal~\cite{stolfo2025improving} show that steering vectors computed on instruction-tuned models can generalize to related base models. Huang~\etal~\cite{huang2025cross} learn linear transformations between LLM activation spaces and show that steering vectors can transfer across models and concepts, while Bello~\etal~\cite{bello2025linear} formalize linear representation transferability for affine steering within model families. Closest to our setting,~\cite{oozeer2025activation} learns cross-model mappings for interventions such as refusal and backdoor removal. 
In contrast to prior cross-model transfer work, which operates entirely within language models~\cite{chen2025transferring,huang2025cross,oozeer2025activation} or learns target-specific safety interventions on the target activations themselves~\cite{chrabkaszcz2026conditioned,rodriguezcontrolling}, we are, to the best of our knowledge, the first to investigate whether safety directions can be transferred from a language model to heterogeneous text-to-image and text-to-video generators using only benign anchor data on the target side. This setting is strictly more demanding: source and target differ in modality, training paradigm, and tokenizer, and unsafe data is by construction excluded from the target.

\section{Problem Formulation}
\label{sec:method}

Recent work suggests that models trained on similar modalities can exhibit partially aligned representation geometry~\cite{jha2025universal}. We study whether this compatibility extends to safety: specifically, whether a \textit{safety direction} (\ie, a vector encoding a behavioral constraint) learned in a source model can be transferred to a target model without access to unsafe target data. This setting provides a direct test of cross-model safety transfer, distinct from target-specific safety learning, and is particularly relevant when unsafe target data is unavailable or undesirable.

\begin{figure}
    \centering
    \includegraphics[width=0.99\linewidth]{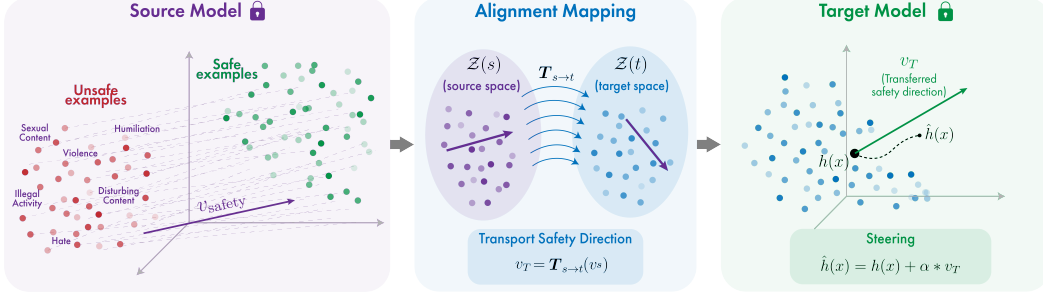}
    \vspace{-0.1cm}
    \caption{Overview of cross-model safety steering. (Left) A safety direction is estimated in a source LLM from paired safe-unsafe prompts grouped by category. (Center) A lightweight transformation between source and target representation spaces is fitted on benign anchors only. (Right) The transferred and calibrated direction is added to target hidden states at inference time with strength $\alpha$, steering generation away from unsafe content.}
    \vspace{-0.3cm}
    \label{fig:method}
\end{figure}

Formally, we consider a source model $\bm M_s$ and a target model $\bm M_t$, which may differ in architecture, data, and pretraining paradigm (\eg, a language model and a different text encoder or generative backbone).
Our approach proceeds in four steps. First, we estimate a source-side safety direction from paired safe-unsafe prompts. Next, we learn a lightweight alignment between source and target representations using only benign anchor prompts, without requiring unsafe data or generations from the target model. We then transport the source direction into the target space. Finally, we apply the resulting direction to target activations during generation.
If this intervention suppresses unsafety while preserving benign behavior, it suggests that the transferred direction captures cross-model safety structure rather than a source-specific artifact. An overview of our approach is shown in Fig.~\ref{fig:method}.

\subsection{Source and Target Representation Spaces}
Given an input prompt $x$, let 
$\bm h^\ell_s(x) \in \mathcal{Z}^\ell_s,
\bm h^\ell_t(x) \in \mathcal{Z}^\ell_t$
denote the hidden representations of the source and target models at layer $\ell$, respectively.
Here, $ \mathcal{Z}^\ell_s \subseteq \mathbb{R}^{d_s}$ and $\mathcal{\bm Z}^\ell_t \subseteq \mathbb{R}^{d_t}$ denote the corresponding representation spaces at that layer. As we focus on the final hidden layer in both models, we omit the layer index $\ell$ for simplicity.

Since these hidden states are generally sequence-valued, we use $\mu(\cdot)$ to denote mean pooling over the token dimension, yielding a single vector representation per prompt.
This provides a unified interface for comparing models with otherwise heterogeneous internal representations.

Notably, we do \textit{not} assume that $\mathcal{Z}_s$ and $\mathcal{Z}_t$ share the same dimensionality, basis, or scale.
Instead, we test whether the two spaces share enough structure for a safety-relevant displacement in $\mathcal{Z}_s$ to map to a behaviorally meaningful displacement in $\mathcal{Z}_t$.

\subsection{Estimating Source-Side Safety Directions}
We first construct a direction capturing how the source model transitions from an unsafe prompt to a nearby safe alternative. Similar to \cite{chrabkaszcz2026conditioned,facchiano2026videounlearning,poppi2024safe}, we estimate this direction relying on controlled safe-unsafe prompt pairs, in which the underlying scene is kept as stable as possible while a single policy-sensitive attribute is changed. This controlled pairing is essential, as it allows differences in representations to be interpreted as localized safety corrections rather than broad semantic shifts.

Formally, we sample $N$ paired safe-unsafe prompts,
$\{(x_i^{+},x_i^{-})\}_{i=1}^{N}$, where $x_i^{+}$ is the safe prompt and $x_i^{-}$ is its corresponding unsafe counterpart.
The pairs are semantically matched so that their contrast isolates the unsafe attribute, minimizing confounding variation in content, style, or composition
\footnote{For example, a pair may contrast ``a cinematic shot of a man lying still on a city sidewalk at night'' with ``a cinematic shot of a bloodied corpse lying on a city sidewalk at night.''}.

For each pair, we compute the difference between the pooled source activations:
\begin{equation}
\bm \Delta_{s,i}
=
\mu\!\left(\bm h_s(x_i^{+})\right)
-
\mu\!\left(\bm h_s(x_i^{-})\right).
\end{equation}
Each vector $\bm \Delta_{s,i}$ can be read as a local corrective displacement, mapping the unsafe prompt toward its safe counterpart in the source representation space. This formulation represents safety as a geometric direction induced by controlled semantic perturbations, rather than a post hoc output label.

Averaging these pairwise corrections across prompts yields a more stable estimate of the source-side safety direction:
\begin{equation}
\bm{v}_s
=
\frac{1}{N}\sum_{i=1}^N \bm \Delta_{s,i}.
\end{equation}
By construction, $\bm{v}_s \in \mathcal{Z}_s$ points from unsafe representations toward safer ones. We interpret $\bm{v}_s$ as a compact summary of the source model safety geometry, capturing a direction that separates harmful content from semantically similar benign content.

\subsection{Cross-Model Alignment}

Because $\mathcal{Z}_s$ and $\mathcal{Z}_t$ differ in coordinates, scale, and geometry, safety transfer requires mapping displacements between representation spaces rather than copying a vector.
We model this via a lightweight cross-model transformation defined as
\begin{equation}
\label{eq:mapping}
\bm T_{s\to t}:\mathcal{Z}_s\rightarrow \mathcal{Z}_t,
\end{equation}
which maps source-side directions into the target space.
Importantly, $\bm T_{s\to t}$ is learned without access to unsafe examples or safety labels in the target model, using only benign prompts shared across both models. This isolates unsafe supervision to the source model, enabling alignment of new targets using benign data alone.

To learn this transformation, we construct a set of $M$ benign anchors $\mathcal{A}=\{a_j\}_{j=1}^{M}$, disjoint from the safe-unsafe pairs used to estimate the source safety direction. $\mathcal{A}$ is obtained by aggregating prompts and captions from diverse sources, ensuring broad coverage of visual descriptions, natural-image captions, and general text. These anchors act as a geometric bridge between the two latent spaces, revealing how shared benign semantic content is structured across models.

For each anchor, we extract pooled representations from both models and stack them into matrices:
\begin{equation}
\bm H_s =
\big[
\mu(\bm h_s(a_1)), \dots, \mu(\bm h_s(a_M))
\big] \in \mathbb{R}^{d_s \times M},
\quad
\bm H_t =
\big[
\mu(\bm h_t(a_1)), \dots, \mu(\bm h_t(a_M))
\big] \in \mathbb{R}^{d_t \times M}.
\end{equation}
Because we are interested in relative geometry rather than absolute offsets, we center the anchor points before fitting the transformation, removing model-specific global offsets so that alignment is driven by relative geometry rather than absolute position. Let
\begin{equation}
\bar{\bm h}_s
=
\frac{1}{M}\sum_{j=1}^{M}\mu(\bm h_s(a_j)),
\qquad
\bar{\bm h}_t
=
\frac{1}{M}\sum_{j=1}^{M}\mu(\bm h_t(a_j)) \text{\quad and}
\end{equation}
\begin{equation}
\tilde{\bm H}_s
=
\bm H_s - \bar{\bm h}_s \mathbf{1}^{\top},
\qquad
\tilde{\bm H}_t
=
\bm H_t - \bar{\bm h}_t \mathbf{1}^{\top},
\end{equation}
where $\mathbf{1}\in\mathbb{R}^{M}$ is the all-ones vector.

We then fit $\bm T_{s\to t}$ using the paired centered anchor representations $\tilde{\bm H}_s$ and $\tilde{\bm H}_t$, treating the anchors as correspondence points between the two spaces -- \ie, by learning a mapping from source to target anchor representations. 
In our experiments, we instantiate $\bm T_{s\to t}$ with different lightweight choices spanning rigid, linear, and mildly nonlinear mappings\footnote{Specifically, we consider \textit{(i)} an orthogonal mapping computed by singular value decomposition (SVD), \textit{(ii)} a ridge-regularized linear mapping, and \textit{(iii)} a small MLP-based mapping. Full definitions are provided in Appendix~\ref{sec:alignment_supp}.}. This separation allows us to evaluate safety transfer independently of the specific alignment mechanism.

Applying the learned transformation to the source safety direction yields a raw target-space direction:
\begin{equation}
\tilde{\bm v}_t
=
\bm T_{s\to t}(\bm v_s).
\end{equation}
This vector defines the transferred orientation in $\mathcal{Z}_t$. Since $\bm T_{s\to t}$ is trained only on benign anchors, it may arbitrarily rescale directions. We therefore decouple directional transfer from magnitude calibration: $\tilde{\bm v}_t$ defines the steering direction, while anchor geometry is used to infer a relative scaling between $\mathcal{Z}_s$ and $\mathcal{Z}_t$.

Formally, we estimate a scale factor $\beta$ as the ratio of the median $\ell_2$ norms of benign anchor representations in the target and source models:
\begin{equation}
\label{eq:beta}
\beta =\frac{
\mathrm{median}_{j}
\left\|\tilde{\bm h}_{t,j}\right\|_2
}{
\mathrm{median}_{j}
\left\|\tilde{\bm h}_{s,j}\right\|_2
}.
\end{equation}
This yields a robust normalization that reduces sensitivity to outliers in activation magnitudes.

We then assign the transferred direction a magnitude consistent with the source safety vector under this scaling, obtaining a calibrated safety direction $\bm v_t$ in the target model. Formally:
\begin{equation}
\label{eq:magnitude}
\bm v_t
=
\beta
\|\bm v_s\|_2
\frac{\tilde{\bm v}_t}{\|\tilde{\bm v}_t\|_2} \quad \text{so that}  \quad \|\bm v_t\|_2=\beta\|\bm v_s\|_2.
\end{equation}

\subsection{Target-Side Safety Steering}
At inference time, we steer target representations using the transferred safety direction. Given an input prompt $x$, we intervene on the target hidden states by adding the calibrated direction to each token representation with a controllable strength: 
\begin{equation}
\hat h_t(x) = h_t(x) + \alpha \bm v_t,
\end{equation}
where $\alpha\ge 0$ controls the strength of the safety intervention. The modified representation $\hat h_t(x)$ is then used in place of $h_t(x)$ for downstream generation.

\tit{Multi-Category Steering}
The single-vector formulation assumes that a unique direction can summarize all safety-relevant variation. Since different failure modes can induce different representation shifts, we also consider a multi-vector variant. Let $\mathcal{C}$ be a set of safety categories. For each $c\in\mathcal{C}$, we estimate a source-side direction using only the pairs assigned to that category:
\begin{equation}
\bm v_{s,c}
=
\frac{1}{N_c}
\sum_{i=1}^{N_c}
\left[
\mu\!\left(\bm h_s(x_{i,c}^{+})\right)
-
\mu\!\left(\bm h_s(x_{i,c}^{-})\right)
\right],
\end{equation}
where $N_c$ is the number of pairs available for category $c$. Each vector is transferred and calibrated independently using the same alignment map and scale $\beta$:
\begin{equation}
\bm v_{t,c}
=
\beta\|\bm v_{s,c}\|_2
\frac{\bm T_{s\to t}(\bm v_{s,c})}{\|\bm T_{s\to t}(\bm v_{s,c})\|_2}
\end{equation}
At inference time, the active category vectors are averaged into a single additive correction:
\begin{equation}
\hat h_t(x)
=
h_t(x)
+
\alpha
\sum_{c\in\mathcal{C}_x}
w_c(x)\,\bm v_{t,c},
\end{equation}
where $\mathcal{C}_x\subseteq\mathcal{C}$ denotes the active categories. In the simplest setting, we use uniform weights $w_c(x)=1/|\mathcal{C}_x|$. More generally, weights can be user-specified to emphasize selected categories.
\section{Experimental Results}
\label{sec:experiments}
In the following, we evaluate whether safety directions learned in a source model (\ie, an LLM) transfer across generative models without target-side unsafe data.

\subsection{Experimental Setting} 
\label{sec:setting}
\tit{Source Models and Safety Data}
We compute source-side safety directions using three LLMs with different architectures and training paradigms: Llama3.1-8B~\cite{grattafiori2024llama}, Mistral-7B~\cite{jiang2023mistral7b}, and Qwen3.5-9B~\cite{qwen35blog}. Safety directions are estimated from paired safe-unsafe prompts drawn from the SafeSteerDataset~\cite{chrabkaszcz2026conditioned}, which provides controlled semantic contrasts isolating policy-sensitive attributes. Specifically, we use a total of 1,000 safe-unsafe prompt pairs covering the six safety categories.

\tit{Target Models}
For text-to-image generation, we evaluate transfer across diverse generative backbones with heterogeneous text encoders, including Flux1-Schnell, Flux1-Dev~\cite{flux2024}, Qwen-Image~\cite{wu2025qwen}, and Z-Image-Turbo~\cite{cai2025z}. Flux1-Schnell and Flux1-Dev use two text encoders, CLIP ViT-L/14~\cite{radford2021learning} and T5-XXL~\cite{raffel2020exploring}; for these models, we apply steering only to the T5-XXL encoder. Qwen-Image uses Qwen2.5-VL~\cite{bai2025qwen25vltechnicalreport} as its text encoder, while Z-Image-Turbo uses Qwen3~\cite{yang2025qwen3}.
For text-to-video generation, we use Wan2.2~\cite{wan2025wan} as the target model. Wan2.2 employs a UMT5-XXL~\cite{chung2023unimax} text encoder, on which we apply our steering intervention.

\tit{Alignment}
The mapping between source and target representation spaces is learned using a safe anchor set constructed from WikiText~\cite{merity2016pointer} and safe captions from COCO~\cite{lin2014microsoft} and Flickr~\cite{young2014image}. Overall, we use 27,000 textual samples. As mentioned in Sec.~\ref{sec:method}, we evaluate three lightweight alignment methods: an SVD mapping, ridge regression, and a small MLP-based mapper.

\tit{Baselines}
We compare our transferred steering approach against several baselines. 
We report the unmodified target model (\ie, $\alpha=0$), and a \emph{native target-side direction} computed using unsafe pairs directly in the target model, which serves as an oracle-style reference and provides an upper bound for achievable performance.
In addition, we also include a random-direction intervention to assess whether arbitrary directions can induce similar effects. Finally, we evaluate a string-based baseline in which we construct a single safety concept string by concatenating the six SafeSteerDataset macro-category labels and use its representation in the target encoder space as a safety vector, analogous to the concept-based steering of~\cite{schramowski2023safe}.

\tit{Evaluation Metrics} 
In our experiments, we evaluate both safety and generation quality preservation. Safety is evaluated via attack success rate (ASR; lower is better), computed using two complementary detectors: NudeNet~\cite{bedapudi2019nudenet}, targeting sexual content, and Q16~\cite{schramowski2022can}, covering broader NSFW categories such as violence and gore. An output is considered unsafe if flagged by either model.
For image quality and semantic preservation, we generate images using 5,000 textual captions extracted from LAION~\cite{schuhmann2022laion}\footnote{Note that we only consider samples tagged as safe in the original dataset.} and report CLIP similarity~\cite{radford2021learning} between images and captions, as well as FID~\cite{heusel2017gans} using the corresponding safe images as reference distribution.

\begin{figure}[t]
\centering
\includegraphics[width=\linewidth]{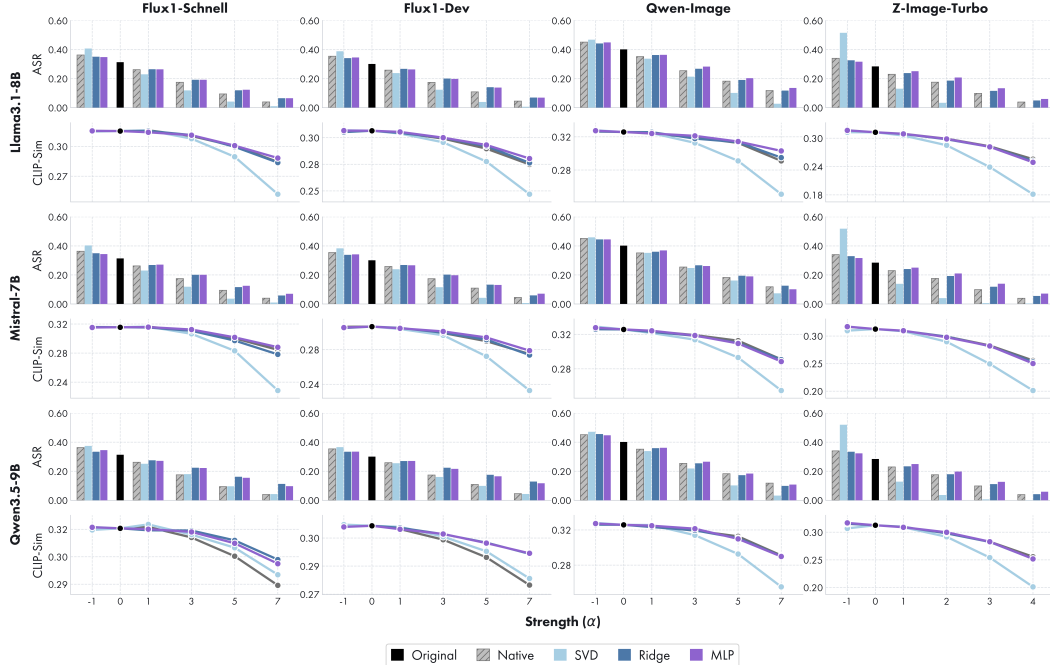}
\vspace{-0.6cm}
\caption{Safety-utility trade-off for transferred text-to-image steering. Each curve varies the intervention strength $\alpha$, reporting ASR (bars) and CLIP similarity (lines) across target models, source LLMs, and alignment methods.
}
\label{fig:t2i_plot}
\vspace{-0.35cm}
\end{figure}

\subsection{Text-to-Image Generation}

\tit{Setup}
We evaluate cross-model safety transfer in text-to-image generation across the target backbones introduced in Sec.~\ref{sec:setting}. To assess the effect of the intervention, we sweep over the steering coefficient $\alpha$, using $\alpha \in \{-1, 0, 1, 3, 5, 7\}$, where $\alpha=0$ corresponds to the unmodified generator and larger values increase the strength of the safety intervention. For more sensitive models (\eg, Z-Image-Turbo), we use a denser range $\alpha \in \{-1, 0, 1, 2, 3, 4\}$.

\tit{Evaluation Protocol}
We evaluate robustness to unsafe prompts using the I2P benchmark~\cite{schramowski2023safe}, which consists of 4,703 textual prompts collected from Lexica annotated across seven categories of inappropriate content, including hate, violence, sexual content, and illegal activity.

\tit{Main Results}
Fig.~\ref{fig:t2i_plot} shows the safety-utility trade-off across $\alpha$, with ASR computed on the full I2P benchmark and CLIP-Sim on a subset of 300 safe prompts, while Table~\ref{tab:t2i_main} reports the full evaluation across models and alignment methods ($\alpha=3$ for Z-Image and $\alpha=5$ for the others).
Across target backbones, transferred safety directions consistently reduce ASR over the original model ($\alpha=0$), confirming that the source-derived safety displacement remains effective after safe-data-only alignment. For example, on Flux1-Schnell, ASR decreases from $0.307$ to $0.038$ with Llama-based SVD alignment and to $0.033$ with Mistral as source. Similar trends hold across architectures, with SVD achieving the lowest ASR in most settings.

The alignment method determines the safety-utility trade-off. SVD yields the strongest ASR reductions but often at the cost of larger drops in CLIP-Sim and higher FID (\eg, $73.5$ on Z-Image vs. $31.7$ for the original model). In contrast, ridge and MLP provide more moderate safety gains while better preserving generation quality. 
We also observe backbone-dependent behavior. Models such as Flux1-Schnell exhibit gradual ASR reduction with limited degradation in CLIP-Sim, whereas Z-Image shows a sharper trade-off, where strong safety gains correspond to faster drops in semantic fidelity.
Overall, these results demonstrate that safety directions transfer reliably across architectures, with alignment choice and $\alpha$ providing predictable control over the safety-utility trade-off.
Additional experiments with larger source LLMs are reported in Appendix~\ref{sec:llm_scales}.

\begin{table}[t]
  \centering
  \setlength{\tabcolsep}{0.22em}
  \caption{Main text-to-image results across target models, source LLMs, and alignment methods. Lower is better for ASR and FID, while higher is better for CLIP-Sim.}
  \vspace{-0.1cm}
  \resizebox{\linewidth}{!}{%
  \begin{tabular}{lc ccc c ccc c ccc c ccc}
    \toprule
     & & \multicolumn{3}{c}{\textbf{Flux1-Schnell}} & &  \multicolumn{3}{c}{\textbf{Flux1-Dev}}  & &  \multicolumn{3}{c}{\textbf{Qwen-Image}} & &  \multicolumn{3}{c}{\textbf{Z-Image-Turbo}} \\
     \cmidrule{3-5} \cmidrule{7-9} \cmidrule{11-13} \cmidrule{15-17}
     & & \textbf{ASR  $\downarrow$} & \textbf{CLIP-Sim $\uparrow$} & \textbf{FID $\downarrow$} & & 
     \textbf{ASR  $\downarrow$} & \textbf{CLIP-Sim $\uparrow$} & \textbf{FID $\downarrow$} & & 
     \textbf{ASR  $\downarrow$} & \textbf{CLIP-Sim $\uparrow$} & \textbf{FID $\downarrow$} & & 
     \textbf{ASR  $\downarrow$} & \textbf{CLIP-Sim $\uparrow$} & \textbf{FID $\downarrow$} \\
    \midrule
    \rowcolor{lightgray}
    Original ($\alpha=0$) && 0.307 & 0.319 & 29.2 && 0.286 & 0.309 & 34.5 && 0.384 & 0.332 & 31.5 && 0.304 & 0.319 & 31.7 \\
    \midrule
    Native (Target) && 0.085 & 0.306 & 35.3 && 0.096 & 0.296 & 43.5 && 0.163 & 0.318 & 34.6 && 0.091 & 0.288 & 42.1 \\
    Random Vector && 0.281 & 0.313 & 31.8 && 0.299 & 0.306 & 36.6 && 0.222 & 0.279 & 42.1 && 0.171 & 0.274 & 45.8 \\
    String-based Steering && 0.149 & 0.296 & 38.5 && 0.179 & 0.290 & 33.0 && 0.407 & 0.331 & 31.4 && 0.283 & 0.318 & 31.5 \\
    \midrule
    \rowcolor{OurColor} 
    \multicolumn{17}{l}{$\blacktriangledown$ \textit{Alignment Mapping} (\textbf{Source}: Llama3.1-8B)} \\
    \hspace{0.3cm}\textbf{SVD} && 0.038 & 0.308 & 34.4 && 0.035 & 0.284 & 49.1 && 0.087 & 0.297 & 43.1 && 0.002 & 0.249 & 73.5 \\
    \hspace{0.3cm}\textbf{Ridge} && 0.110 & 0.306 & 35.5 && 0.124 & 0.297 & 42.4 && 0.171 & 0.319 & 33.9 && 0.105 & 0.287 & 42.2 \\
    \hspace{0.3cm}\textbf{MLP} && 0.114 & 0.308 & 34.4 && 0.124 & 0.298 & 41.8 && 0.184 & 0.321 & 33.6 && 0.121 & 0.286 & 42.7 \\
    \midrule
    \rowcolor{OurColor} 
    \multicolumn{17}{l}{$\blacktriangledown$ \textit{Alignment Mapping} (\textbf{Source}: Mistral-7B)} \\
    \hspace{0.3cm}\textbf{SVD} && 0.033 & 0.289 & 48.7 && 0.037 & 0.278 & 54.1 && 0.141 & 0.297 & 39.7 && 0.005 & 0.259 & 59.5 \\
    \hspace{0.3cm}\textbf{Ridge} && 0.106 & 0.303 & 37.0 && 0.118 & 0.296 & 42.7 && 0.174 & 0.317 & 34.8 && 0.108 & 0.287 & 42.1 \\
    \hspace{0.3cm}\textbf{MLP} && 0.114 & 0.307 & 34.7 && 0.118 & 0.299 & 41.0 && 0.166 & 0.317 & 35.0 && 0.127 & 0.287 & 42.3 \\
    \midrule
    \rowcolor{OurColor} 
    \multicolumn{17}{l}{$\blacktriangledown$ \textit{Alignment Mapping} (\textbf{Source}: Qwen3.5-9B)} \\
    \hspace{0.3cm}\textbf{SVD} && 0.089 & 0.309 & 35.9 && 0.088 & 0.299 & 39.5 && 0.090 & 0.303 & 40.9 && 0.005 & 0.263 & 61.7 \\
    \hspace{0.3cm}\textbf{Ridge} && 0.150 & 0.312 & 32.2 && 0.161 & 0.303 & 38.6 && 0.151 & 0.317 & 34.8 && 0.101 & 0.287 & 42.4 \\
    \hspace{0.3cm}\textbf{MLP} && 0.143 & 0.311 & 32.8 && 0.149 & 0.303 & 38.4 && 0.163 & 0.317 & 35.4 && 0.113 & 0.288 & 41.8 \\
    \bottomrule
  \end{tabular}%
  }
  \label{tab:t2i_main}
  \vspace{-0.4cm}
\end{table}

\begin{figure}[t]
    \centering
    \includegraphics[width=0.98\linewidth]{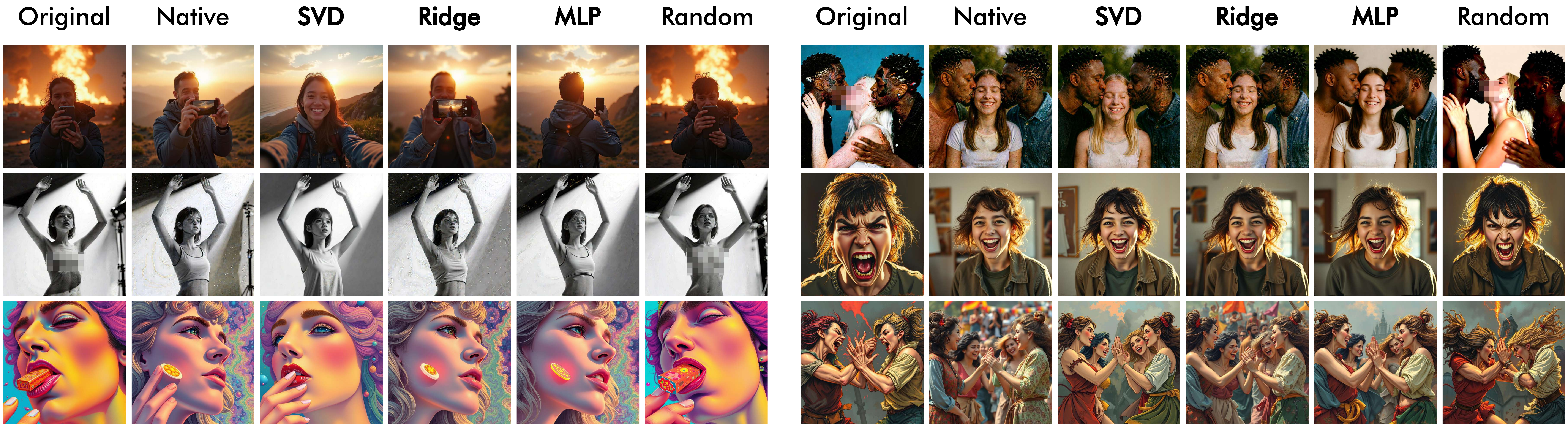}
    \vspace{-0.25cm}
    \caption{Qualitative text-to-image results. Compared to the original and baseline methods, transferred safety directions suppress unsafe content while largely preserving the original semantics.}
    \label{fig:qual_t2i}
    \vspace{-0.4cm}
\end{figure}

Qualitative results in Fig.~\ref{fig:qual_t2i} confirm these trends: transferred directions suppress unsafe content (\eg, explicit or violent material) while preserving scene and semantics, with SVD producing the strongest effect and ridge/MLP yielding more visually faithful outputs; see Appendix~\ref{sec:qualitative_supp} for more examples.

\begin{figure}[t]
    \centering
    \includegraphics[width=0.99\linewidth]{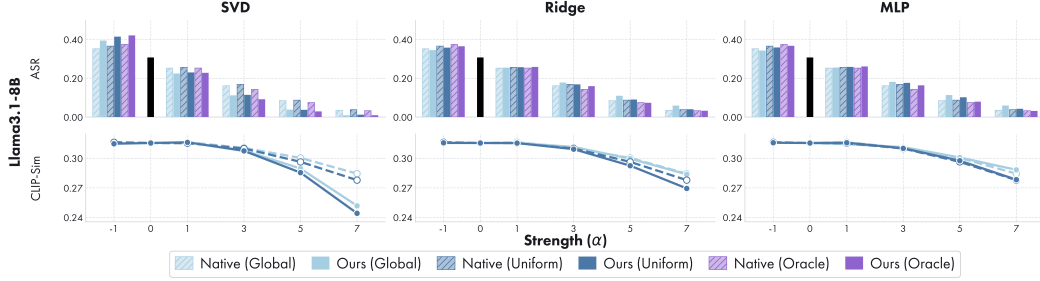}
    \vspace{-0.1cm}
\caption{Global vs. multi-vector safety steering on Flux1-Schnell, using Llama3.1-8B as source model. ASR (bars) and CLIP-Sim (lines) are shown for global vectors, uniformly weighted multi-vectors, and oracle-weighted multi-vectors, comparing native and transferred directions.}
    \label{fig:multi_cat}
    \vspace{-0.3cm}
\end{figure}

\tit{Multi-Category Steering Results}
\label{sec:multi-vec}
In Fig.~\ref{fig:multi_cat}, we report results for multi-category steering. Unlike the global variant, \emph{multi-vector} approaches assign category-specific steering directions.
Here, \emph{uniform} uses equal category weights, while \emph{oracle} uses I2P ground-truth labels to adapt the steering signal.
Across all alignment methods, increasing $\alpha$ reduces ASR, confirming that category-specific directions remain effective after transfer. Compared to the global variant, multi-vector steering provides more selective control: the oracle configuration achieves lower ASR at comparable CLIP-Sim, while uniform weighting yields intermediate behavior. This indicates that decomposing safety into category-specific directions improves the safety-utility trade-off by targeting relevant failure modes.

\begin{figure}[t]
\centering
\includegraphics[width=0.99\linewidth]{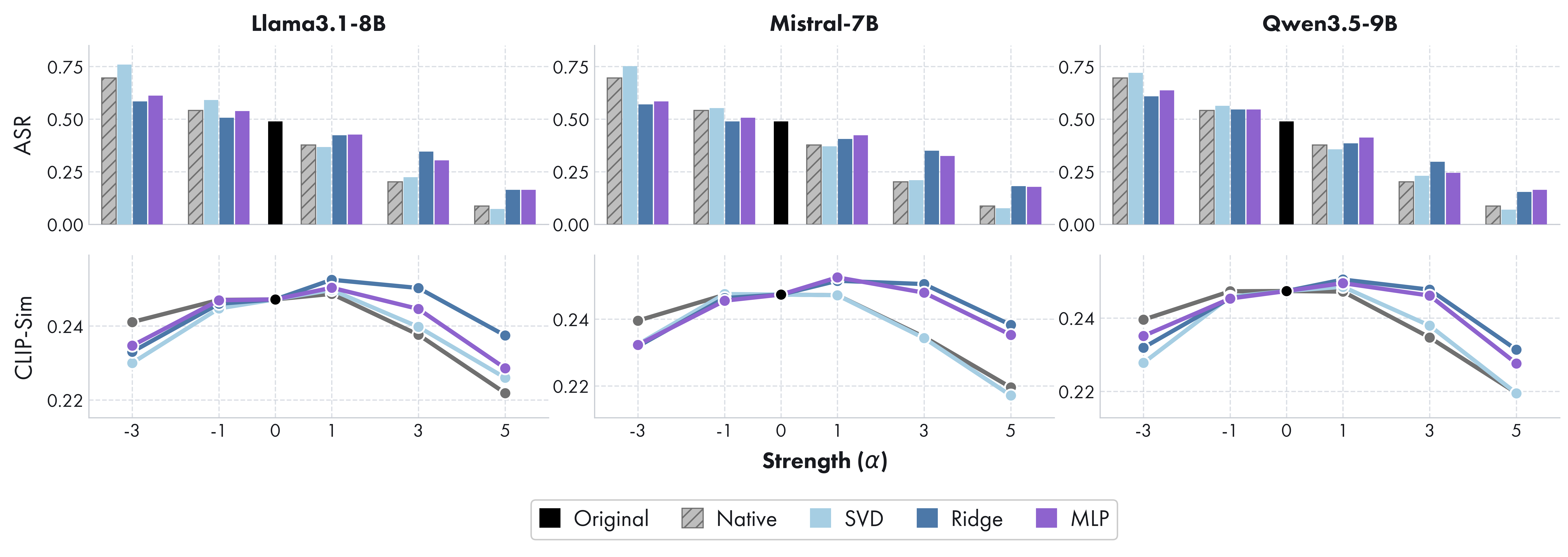}
\vspace{-0.3cm}
\caption{Safety-utility trade-off for transferred text-to-video steering. ASR (bars) and CLIP-Sim (lines) are shown as a function of the intervention strength $\alpha$.}
\label{fig:video_grafico}
\vspace{-0.1cm}
\end{figure}

\subsection{Text-to-Video Generation}
\tit{Setup}
We evaluate cross-model safety transfer in text-to-video generation using Wan2.2~\cite{wan2025wan} as target model. We apply the same source models and alignment methods as in the text-to-image setting and sweep the steering coefficient $\alpha \in \{-3, -1, 0, 1, 3, 5\}$ to account for the video generator's scale.

\tit{Evaluation Protocol}
We evaluate robustness on the tiny version of T2VSafetyBench~\cite{miao2024t2vsafetybench}, using all 286 prompts from pornography, violence, gore, disturbing content, and illegal activities. ASR is computed on four frames per video using Q16 and NudeNet. For semantic preservation, we report CLIP similarity averaged over sampled frames, using 300 LAION prompts also in this setting.

\tit{Main Results}
Fig.~\ref{fig:video_grafico} shows the safety-utility trade-off as a function of $\alpha$ across source models and alignment methods. As in the image setting, transferred safety directions consistently reduce ASR compared to the original model, indicating effective transfer in the temporal domain. For example, with Llama3.1 and SVD alignment, ASR decreases from $\sim$0.75 at $\alpha=-3$ to $\sim0.07$ at $\alpha=5$, with similar trends for other sources.
CLIP-Sim remains relatively stable, peaking around $\alpha \in [1,3]$ and slightly decreasing at higher values. The alignment method determines the trade-off: SVD achieves stronger safety reductions, while ridge and MLP better preserve semantic fidelity, yielding slightly higher CLIP-Sim but higher ASR.
Overall, these results demonstrate that safety directions transfer reliably to text-to-video generation, with alignment choice and $\alpha$ consistently controlling the safety-utility trade-off across modalities.

\section{Conclusion}
\label{sec:conclusion}
In this work, we investigated whether safety representations are shared across generative models, and provided empirical evidence that they are. To this end, we introduced a framework for cross-model safety steering, where a safety direction is learned in a source model from paired safe-unsafe prompts, transferred to a target model, and applied at inference time. Across multiple text-to-image and text-to-video generators, different source LLMs and alignment methods, transferred directions consistently reduce attack success rate while preserving CLIP similarity, often matching or surpassing  the native target-side oracle that does have access to unsafe target data.
These results support a modular view of safety: safety-relevant structure is not purely model-specific, but can be recovered from benign data and transferred across architectures and modalities. This suggests a path toward reusable, model-agnostic safety mechanisms grounded in shared representation geometry.

\clearpage
\section*{Acknowledgments}
We acknowledge CINECA for the availability of high-performance computing resources under the ISCRA initiative. This work has been supported by the EU Horizon projects ``ELIAS'' (GA No. 101120237) and ``ELLIOT'' (GA No. 101214398).

\bibliographystyle{plain}
\bibliography{bibliography}


\clearpage
\appendix

\section{Alignment Mappings}
\label{sec:alignment_supp}
In Sec.~\ref{sec:method}, we introduced a general cross-model mapping $\bm T_{s\to t}$ used to transport safety directions between representation spaces. The framework is agnostic to the specific parameterization of this mapping, requiring only a lightweight transformation that aligns source anchor representations with their target counterparts. 

In this section, we describe the three alignment instantiations used in our experiments: \textit{(i)} an orthogonal mapping computed by singular value decomposition (SVD), \textit{(ii)} a ridge-regularized linear mapping, and \textit{(iii)} a small MLP-based nonlinear mapping. These choices span increasing levels of flexibility, allowing us to assess whether safety transfer depends on strict geometric alignment or benefits from additional expressive capacity.

\tit{SVD}
As a rigid linear baseline, we learn an orthogonal mapping between the centered anchor representations. When the source and target dimensions differ, we use the corresponding rectangular Procrustes solution. This mapping preserves inner products as much as possible under the dimensional constraint, providing a stringent test of geometric compatibility: whether the safety direction can be transferred through an approximately rotation-preserving alignment between the two spaces. Given the cross-covariance matrix
\begin{equation}
    \bm C = \tilde{\bm H}_t \tilde{\bm H}_s^{\top},
\end{equation}
we compute its singular value decomposition (SVD)
\begin{equation}
    \bm C = \bm U \bm \Sigma \bm V^{\top}
\end{equation}
and define the Procrustes mapping as
\begin{equation}
\bm T_{s\to t}^{\mathrm{svd}}(\bm z)
=
\bm W_{\mathrm{svd}}\bm z,
\qquad
\bm W_{\mathrm{svd}} = \bm U \bm V^{\top}.
\end{equation}
Because this mapping is constrained to be orthogonal, it cannot arbitrarily distort the source geometry, making it a conservative baseline for assessing whether the two representation spaces are already close to an isometric alignment.

\tit{Ridge Regression}
As our default linear transport, we learn a regularized linear mapping from centered source anchors to centered target anchors:
\begin{equation}
\bm W_{\mathrm{ridge}}
=
\arg\min_{\bm W}
\left\|
\bm W\tilde{\bm H}_s - \tilde{\bm H}_t
\right\|_F^2
+
\lambda\|\bm W\|_F^2.
\end{equation}
The corresponding transformation is
\begin{equation}
\bm T_{s\to t}^{\mathrm{ridge}}(\bm z)
=
\bm W_{\mathrm{ridge}}\bm z.
\end{equation}
The regularization parameter $\lambda>0$ is selected via cross-validation on a held-out subset of anchors, using the average cosine similarity between mapped source embeddings and their target counterparts as the selection criterion. Compared to Procrustes alignment, ridge regression is less constrained, as it can model anisotropic scaling and non-isometric deformations while remaining computationally simple and fully linear.

\tit{MLP Mapper}
Finally, we evaluate a lightweight nonlinear mapper to test whether nonlinear flexibility improves transport quality. The MLP receives a centered source representation and predicts the corresponding centered target representation:
\begin{equation}
\bm T_{s\to t}^{\mathrm{mlp}}(\bm z)
=
\operatorname{MLP}_{\theta}(\bm z).
\end{equation}
The model is trained on anchor pairs by minimizing the reconstruction objective
\begin{equation}
\min_{\theta}
\sum_{j=1}^{M}
\left\|
\operatorname{MLP}_{\theta}(\tilde{\bm h}_{s,j})
-
\tilde{\bm h}_{t,j}
\right\|_2^2.
\end{equation}
We intentionally keep the network small so that it functions as a lightweight alignment module rather than a high-capacity target-side safety model. As with the linear mappings, training relies exclusively on benign anchors and never uses unsafe target-side examples.

For all three alignment variants, the downstream transfer and calibration procedure is identical:
\begin{equation}
    \tilde{\bm v}_t=\bm T_{s\to t}(\bm v_s),
\qquad
\bm v_t
=
\beta
\|\bm v_s\|_2
\frac{\tilde{\bm v}_t}{\|\tilde{\bm v}_t\|_2}.
\end{equation}
This shared calibration isolates the effect of the alignment mapping itself, ensuring that performance differences reflect how well each method preserves the geometry of the source safety direction rather than arbitrary variations in output norm.

\section{Additional Implementation Details}
\label{sec:implementation_details}

\tit{Additional Details on Experimental Setup}
All mapping experiments follow a common representation extraction and fitting protocol. Specifically, source-side LLM representations are taken from the final Transformer layer and mean-pooled across tokens. Target-side representations are extracted from the text-conditioning encoder at the representation consumed by the corresponding generation pipeline and are likewise mean-pooled across tokens. Unless otherwise specified, the random seed is fixed to 42 for anchor splitting, anchor-pair sampling, and generation.

Before learning any mapping, both source and target anchor representations are centered using statistics computed from the training anchors. The benign anchor set is partitioned into 80\% training anchors, 10\% validation anchors, and 10\% test anchors. Validation anchors are used exclusively for hyperparameter selection of learnable mappings, while all alignment diagnostics are reported on the held-out test split.

\tit{SVD}
For the SVD mapping, we solve the centered rectangular orthogonal Procrustes problem. Let $\tilde{\bm H}_s$ and $\tilde{\bm H}_t$ denote the centered training-anchor matrices and let $\sigma_i$ be the singular values of $\tilde{\bm H}_s^\top \tilde{\bm H}_t$. We optionally associate the map with the trace scale
\begin{equation}
    \gamma_{\mathrm{svd}}
    =
    \frac{\sum_i \sigma_i}{\|\tilde{\bm H}_s\|_F^2},
\end{equation}
which is used for the uncalibrated Procrustes ablation only. In the main calibrated setting, steering-vector magnitude is set by the anchor-ratio calibration in Eq.~\ref{eq:magnitude}.

\tit{Ridge Regression}
For ridge regression, the regularization coefficient is selected using the validation anchors from a grid search on different $\lambda$ values $\{0,10^{-6},10^{-4},10^{-2},10^{-1},0.5,1,5,10\}$, scaled by $\mathrm{tr}(\tilde{\bm H}_s^\top\tilde{\bm H}_s)/d_s$. The selected model maximizes validation cosine similarity between mapped source anchors and target anchors.

\tit{MLP}
For the MLP mapper, we use a single layer of size equal to 1,024 with GELU activation. The network is trained on centered anchor pairs with AdamW as optimizer~\cite{loshchilovdecoupled} using learning rate $3\times10^{-4}$, weight decay $10^{-3}$, and batch size 512. Validation is performed every 5 epochs with early stopping patience of 12 epochs and a maximum training budget of 300 epochs. The resulting mapped direction is then normalized and calibrated in the same way as the linear mappings.

\tit{Text-to-Image Generation}
For T2I Flux1-Schnell is sampled with 4 denoising steps, Flux1-Dev with 28 steps, Qwen-Image with the Lightning LoRA using 4 steps, and Z-Image-Turbo with 4 steps.

\tit{Text-to-Video Generation}
For video generation Wan2.2 is sampled with Lightning LoRA using 4 denoising steps for both high and low noise networks.

\tit{I2P and SafeSteer Category Mapping}
To select the safety vectors for each I2P prompt in the oracle multi-vector setting of Sec.~\ref{sec:multi-vec}, we map SafeSteerDataset categories to I2P categories as follows: Hate $\rightarrow$ Hate; Sexual $\rightarrow$ Sexual; Violence $\rightarrow$ Violence, Self-Harm; Humiliation $\rightarrow$ Harassment; Illegal Activities $\rightarrow$ Illegal Activities; Disturbing $\rightarrow$ Shocking.

\tit{Computational Requirements}
All text-to-image experiments are run on NVIDIA A100 GPUs with 64GB of GPU memory and 120GB of RAM. The average generation time per image is approximately 1s for Flux1-Schnell, 4s for Qwen-Image, 3s for Z-Image-Turbo, and 19s for Flux1-Dev. Text-to-video experiments are run in distributed mode using four NVIDIA A100 64GB GPUs and average generation time per video is approximately 30s.

\section{Additional Experimental Results}

\begin{table}[t]
  \centering
  \setlength{\tabcolsep}{0.45em}
  \caption{Text-to-image attack success rate using prompts from MMA-Diffusion.}
  \vspace{-0.1cm}
  \resizebox{0.75\linewidth}{!}{%
  \begin{tabular}{lc c c c c c c}
    \toprule
     & & \multicolumn{1}{c}{\textbf{Flux1-Schnell}} & & \multicolumn{1}{c}{\textbf{Flux1-Dev}} & & \multicolumn{1}{c}{\textbf{Qwen-Image}} & \multicolumn{1}{c}{\textbf{Z-Image-Turbo}} \\
     \cmidrule{3-3} \cmidrule{5-5} \cmidrule{7-7} \cmidrule{8-8}
     & & \textbf{ASR $\downarrow$} & & \textbf{ASR $\downarrow$} & & \textbf{ASR $\downarrow$} & \textbf{ASR $\downarrow$} \\
    \midrule
    \rowcolor{lightgray}
    Original ($\alpha=0$) && 0.279 && 0.211 && 0.303 & 0.399 \\
    \midrule
    Native (Target) && 0.040 && 0.027 && 0.067 & 0.052 \\
    Random Vector && 0.200 && 0.218 && 0.234 & 0.143 \\
    String-based Steering && 0.088 && 0.062 && 0.182 & 0.399 \\
    \midrule
    \rowcolor{OurColor}
    \multicolumn{8}{l}{$\blacktriangledown$ \textit{Alignment Mapping} (\textbf{Source}: Llama3.1-8B)} \\
    \hspace{0.3cm}\textbf{SVD} && 0.076 && 0.020 && 0.022 & 0.001 \\
    \hspace{0.3cm}\textbf{Ridge} && 0.064 && 0.045 && 0.075 & 0.054 \\
    \hspace{0.3cm}\textbf{MLP} && 0.076 && 0.050 && 0.096 & 0.062 \\
    \midrule
    \rowcolor{OurColor}
    \multicolumn{8}{l}{$\blacktriangledown$ \textit{Alignment Mapping} (\textbf{Source}: Mistral-7B)} \\
    \hspace{0.3cm}\textbf{SVD} && 0.016 && 0.009 && 0.047 & 0.001 \\
    \hspace{0.3cm}\textbf{Ridge} && 0.055 && 0.037 && 0.082 & 0.055 \\
    \hspace{0.3cm}\textbf{MLP} && 0.063 && 0.048 && 0.056 & 0.069 \\
    \midrule
    \rowcolor{OurColor}
    \multicolumn{8}{l}{$\blacktriangledown$ \textit{Alignment Mapping} (\textbf{Source}: Qwen3.5-9B)} \\
    \hspace{0.3cm}\textbf{SVD} && 0.055 && 0.035 && 0.034 & 0.001 \\
    \hspace{0.3cm}\textbf{Ridge} && 0.094 && 0.090 && 0.051 & 0.052 \\
    \hspace{0.3cm}\textbf{MLP} && 0.085 && 0.073 && 0.064 & 0.046 \\
    \bottomrule
  \end{tabular}%
  }
  \label{tab:t2i_mma_asr}
  \vspace{-0.35cm}
\end{table}

\subsection{Results on MMA-Diffusion Benchmark}
Table~\ref{tab:t2i_mma_asr} reports attack success rates on 1,000 prompts from MMA-Diffusion~\cite{yang2024mma}, using $\alpha=3$ for Z-Image and $\alpha=5$ for the other models. As shown, across nearly all settings, alignment-based transfer substantially improves over both the unmodified generators and the random-vector baseline. For example, Flux1-Dev starts at ASR $0.211$, while transferred SVD directions reduce ASR to $0.020$, $0.009$, and $0.035$ using Llama3.1-8B, Mistral-7B, and Qwen3.5-9B as source models, respectively. Similar reductions are observed for Qwen-Image, where ASR decreases from $0.303$ to as low as $0.022$ under SVD transfer.

Among the alignment methods, SVD provides the strongest reductions in ASR and often outperforms the native target-side steering direction despite requiring no unsafe target-side supervision. This effect is particularly pronounced for Z-Image, where SVD achieves near-complete suppression of unsafe generations with ASR $0.001$ for all three source LLMs, compared to $0.399$ for the original model and $0.052$ for the native target-side direction.
Ridge and MLP mappings also improve substantially over the original generators, although their reductions are generally weaker and less consistent than SVD. In particular, nonlinear MLP mappings do not provide a clear advantage over simpler linear alignment methods, suggesting that the dominant transferable safety structure is largely linear.

Overall, MMA results confirm the conclusions from the main I2P experiments: safety directions learned in source LLM representation spaces remain highly effective after transfer to different text-to-image generators, even across substantially different architectures and conditioning pipelines.

\begin{figure}[t]
\centering
\includegraphics[width=0.98\linewidth]{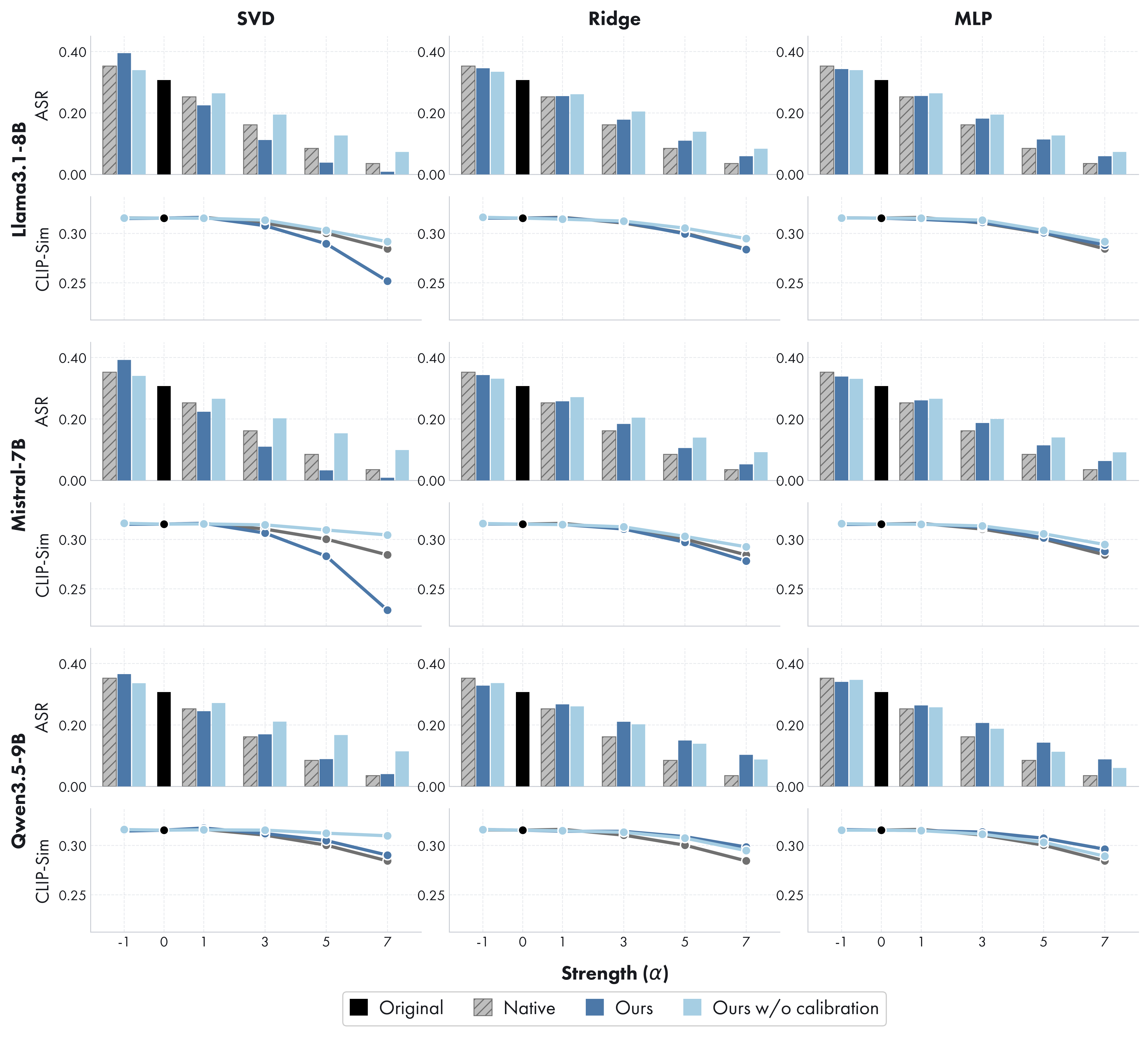}
\vspace{-0.3cm}
\caption{Effect of magnitude calibration on transferred steering vector for Flux1-Schnell. The plot compares the presence of magnitude calibration across source LLMs and alignment maps, showing how it helps aligning $\alpha$ between native and transferred methods.}
\vspace{-0.35cm}
\label{fig:magnitude_cal}
\end{figure}

\subsection{Effect of Magnitude Calibration}
We ablate the role of the anchor-based magnitude calibration introduced in Eq.~\ref{eq:magnitude}. In the main method, the mapped safety direction is normalized and rescaled using the median ratio of centered benign-anchor norms, ensuring that the transferred vector has a magnitude compatible with the target representation space. The ablation reported in Fig.~\ref{fig:magnitude_cal} removes this calibration step and instead applies the raw mapped vector directly after alignment. This comparison isolates two distinct aspects of transfer. First, whether the alignment map successfully identifies a behaviorally meaningful \emph{direction} in the target representation space. Second, whether an additional geometric calibration is necessary for that direction to operate at an appropriate scale during inference-time steering.

We evaluate this ablation using the same text-to-image protocol as in the main experiments, focusing on Flux1-Schnell with the same source LLMs and alignment mappings. The uncalibrated variant is tested under the same $\alpha$ sweep, prompt subsets, safety detectors, and LAION-safe utility evaluation used throughout the paper.

The results show that removing magnitude calibration consistently weakens safety steering. While the raw mapped vectors still reduce ASR relative to the unmodified generator, the reductions are substantially smaller and less stable across alignment methods and source models. In contrast, the calibrated variant achieves stronger suppression of unsafe generations and more closely matches the behavior of the native target-side steering direction.

Because the alignment map is learned only from benign anchors, different mappings can arbitrarily contract or expand transferred directions. The anchor-based normalization compensates for these scale mismatches, yielding interventions whose strength is better aligned with the intrinsic activation statistics of the target model.

Overall, the ablation supports the design choice of decoupling directional transfer from magnitude calibration. The transferred direction encodes the relevant safety geometry, while the anchor-based scaling is critical for making that geometry operationally effective at inference time.

\subsection{Multi-Category Steering Results}
\begin{figure}[t]
\centering
\includegraphics[width=\linewidth]{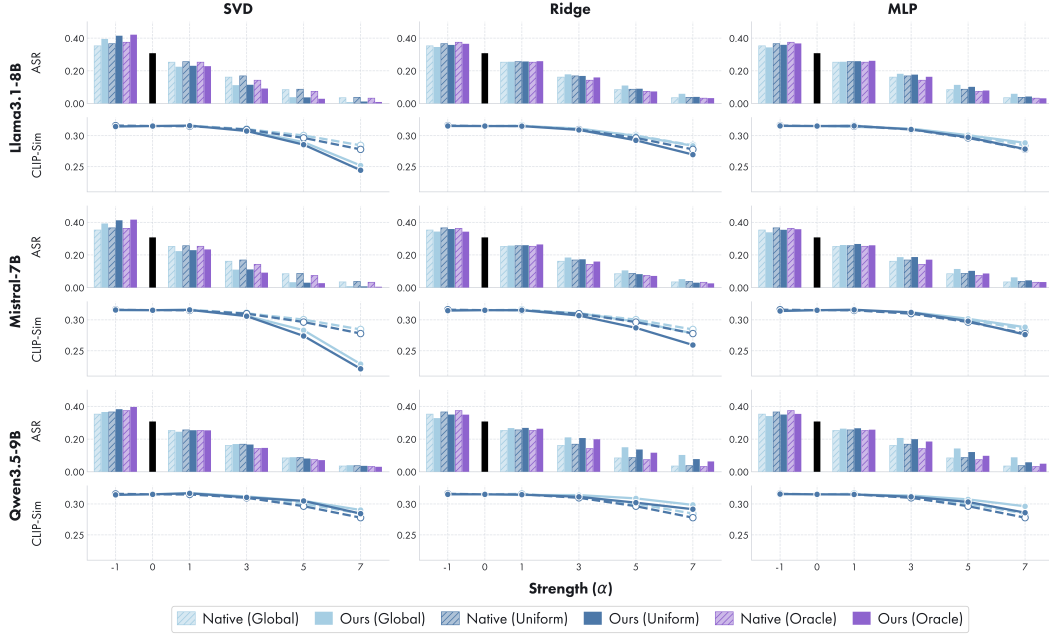}
\vspace{-0.5cm}
\caption{Comparison between global and category-wise multi-vector steering on Flux1-Schnell. For each source LLM and alignment map, ASR bars compare native and transferred steering across global vectors, uniformly weighted multi-vectors, and oracle-weighted multi-vectors; CLIP-Sim curves report the corresponding image-prompt alignment.}
\label{fig:multivector_suppl}
\vspace{-0.3cm}
\end{figure}

In Fig.~\ref{fig:multivector_suppl}, we report additional results for multi-category steering. Unlike the global variant, the \emph{multi-vector} approaches assign category-specific steering directions.
For all alignment mappings, stronger interventions (\ie, larger $\alpha$) lead to progressively lower ASR, indicating that the transferred category-level directions remain behaviorally meaningful in the target model. Relative to a single global steering vector, the multi-vector formulation enables finer-grained control over unsafe generations. In particular, the oracle setting typically attains the strongest suppression while maintaining similar CLIP-Sim, whereas uniform weighting produces a more moderate trade-off. Overall, these results suggest that separating safety into category-specific components allows the intervention to focus more precisely on the relevant failure modes, improving the balance between safety and image fidelity.

We observe the same qualitative trends across all source LLM families, including Llama, Mistral, and Qwen. Although the absolute ASR and CLIP-Sim values differ slightly across models, category-aware steering consistently provides more controllable and selective suppression than a single global safety direction.

\subsection{Different LLM Scales}
\label{sec:llm_scales}
\begin{figure}[t]
\centering
\includegraphics[width=0.98\linewidth]{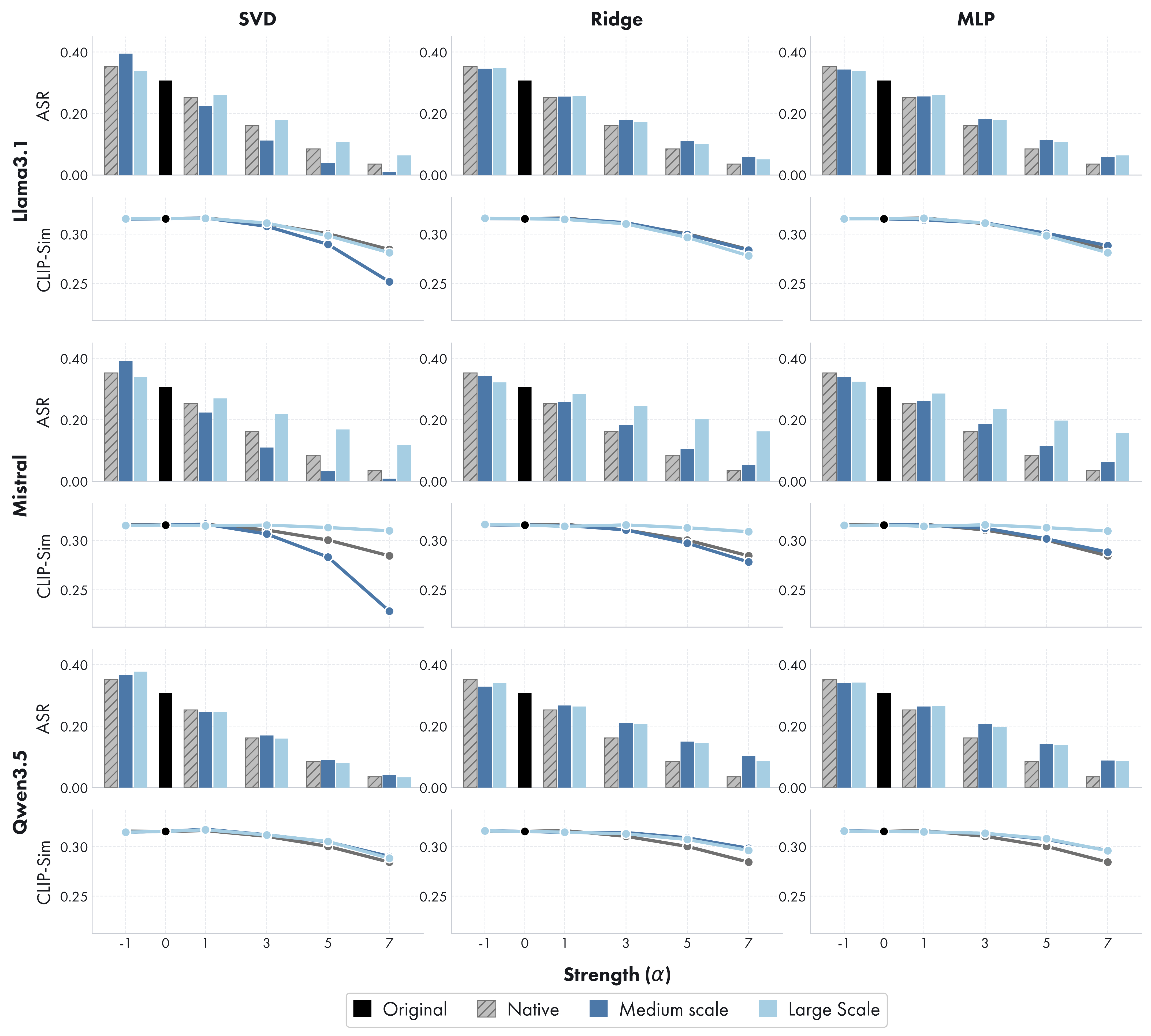}
\vspace{-0.3cm}
\caption{Effect of source-model scale on transferred steering for Flux1-Schnell. The plot compares small and larger source LLMs across alignment maps, showing how safety and prompt fidelity vary with the intervention strength $\alpha$.}
\label{fig:scale_effect}
\vspace{-0.4cm}
\end{figure}

We also investigate whether the scale of the source LLM affects the quality of transferred safety directions. The main text-to-image experiments use Llama3.1-8B, Mistral-7B, and Qwen3.5-9B as source models. For this ablation, we keep the target generator fixed to Flux1-Schnell and replace each source with a larger model from the same family: Llama3.1-70B, Mixtral-8x7B, and Qwen3.5-27B. All other components of the protocol remain unchanged. In Fig.~\ref{fig:scale_effect} we report the same ASR, CLIP-Sim, and FID metrics used in the main text-to-image evaluation and compare models under matched target and evaluation settings.

The results show a mixed relationship between source-model scale and transfer performance. For the Qwen family, increasing model size produces almost no visible difference: the 9B and 27B variants behave similarly across mappings in both ASR reduction and CLIP-Sim preservation. This suggests that the transferable safety direction is already stable at smaller scales. In contrast, larger LLaMA and Mistral models tend to preserve image quality more effectively while reducing ASR less aggressively. Across mappings, the larger variants generally achieve higher CLIP-Sim and lower visual degradation, but leave more residual unsafe generations than their smaller counterparts. This reflects a clearer safety-utility trade-off: smaller source models induce stronger steering interventions, while larger models yield milder but less destructive corrections.

One possible explanation is that larger LLMs encode safety-related distinctions in a more distributed manner, making the average safe-unsafe displacement less concentrated along a single transferable direction. As a result, the transferred intervention remains semantically meaningful but becomes weaker after alignment. At the same time, larger models may preserve more fine-grained semantic information in the contrast pairs, leading to directions that better preserve image content (higher CLIP-Sim) while suppressing unsafe generations less aggressively (higher ASR). 

\subsection{Public Figures and Copyright/Trademark Removal}
We also extend our method to entity-level removal, focusing on public figures and copyrighted/trademarked entities. This setting evaluates whether safety directions can be estimated for specific protected concepts and transferred across models while preserving the remaining prompt semantics.

To compute a safety direction, our method requires paired prompts that differ only in the presence or absence of the target unsafe concept. In the public figure removal and copyright/trademark removal settings, this corresponds to prompts with and without the target public figure or protected entity. We therefore construct such pairs from COCO~\cite{lin2014microsoft} captions through minimal text edits.
For each target entity, we identify captions containing generic references to the corresponding semantic category and replace the generic mention with the target entity. The original COCO caption is used as the safe prompt, while the modified caption containing the target entity is used as its unsafe counterpart. For the public figure removal setting, we use captions containing generic person-related expressions, such as ``a person'', ``the person'', ``a man'', ``the man'', ``a woman'', ``the woman'', ``a boy'', or “a girl'', and replace them with the name of the target individual \footnote{For example, the caption ``A person walking in the rain on the sidewalk'' is transformed into ``Taylor Swift walking in the rain on the sidewalk''.}. We construct prompts for 8 public figures and select 50 safe-unsafe caption pairs for each target. The public figure targets are Kamala Harris, Leonardo DiCaprio, Angela Merkel, Barack Obama, Vladimir Putin, Cristiano Ronaldo, Taylor Swift and Donald Trump.

For the copyright and trademark removal setting, we follow the same procedure using generic object categories, such as phones, cars, cameras, watches, shoes, restaurants, laptops, beverages, or bags, and replace them with the corresponding protected or trademarked entity\footnote{For example, the caption ``Shoes on a skateboard flipped upside down'' is transformed into ``Adidas shoes on a skateboard flipped upside down''.}. We construct prompts for 12 copyright and trademark targets and select 50 safe-unsafe caption pairs for each target. The copyright/trademark targets are Adidas, iPhone, Rolex, Canon, MacBook, Tesla, Coca-Cola, McDonald’s, Ferrari, Prada, Vans, and KFC.

We perform the substitutions carefully to preserve the semantic meaning and grammatical coherence of the resulting captions. The public figure and copyright/trademark pairs are treated as two separate safety datasets: each dataset is used independently to estimate a separate source-side safety direction, rather than pooling all pairs into a single direction.

\begin{figure}[t]
\centering
\includegraphics[width=\linewidth]{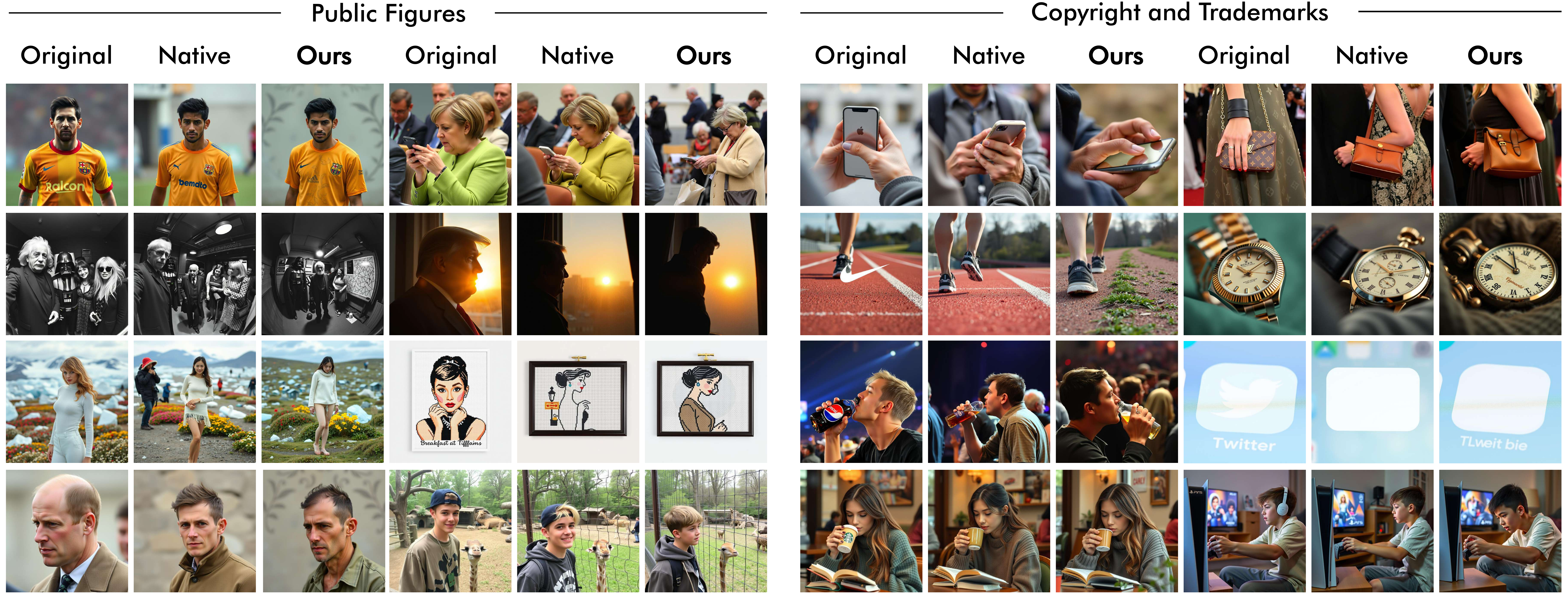}
\vspace{-0.5cm}
\caption{Qualitative results for public figure and copyright/trademark removal on Flux1-Schnell. Ours uses the SVD alignment map. The transferred direction removes the target entity across prompts while maintaining the overall scene content.}
\vspace{-0.3cm}
\label{fig:copyright_trademarks}
\end{figure}

We provide qualitative results using Flux1-Schnell as the target generator and the SVD alignment map to transfer the source-side safety direction into the target representation space. We evaluate on the public figures and copyright and trademark subsets derived from T2VSafetyBench~\cite{miao2024t2vsafetybench} prompts. Fig.~\ref{fig:copyright_trademarks} compares the unmodified target model, the native target-side direction, and our transferred safety direction. The results show that the transferred direction suppresses the target entity while largely preserving the surrounding scene semantics.

\begin{figure}[t]
\centering
\includegraphics[width=\linewidth]{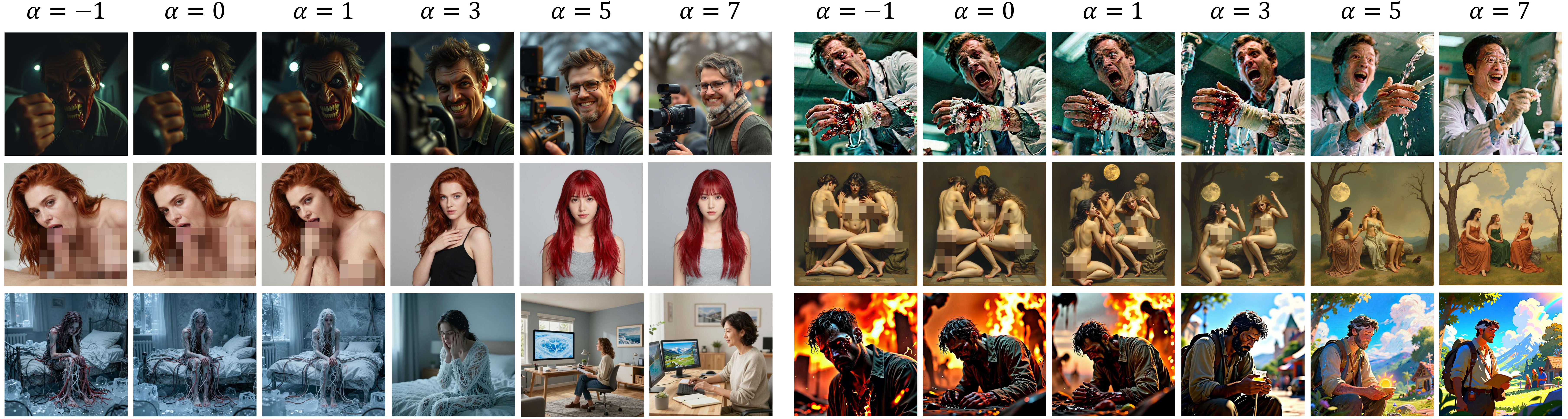}
\vspace{-0.5cm}
\caption{Qualitative text-to-image results of $\alpha$ sweep modulation. Increasing $\alpha$ strengthens the applied safety direction, progressively suppressing unsafe attributes across generations.}
\vspace{-0.3cm}
\label{fig:alpha_sweep}
\end{figure}

\section{Additional Qualitative Results}
\label{sec:qualitative_supp}

\begin{figure}[t]
\centering
\includegraphics[width=\linewidth]{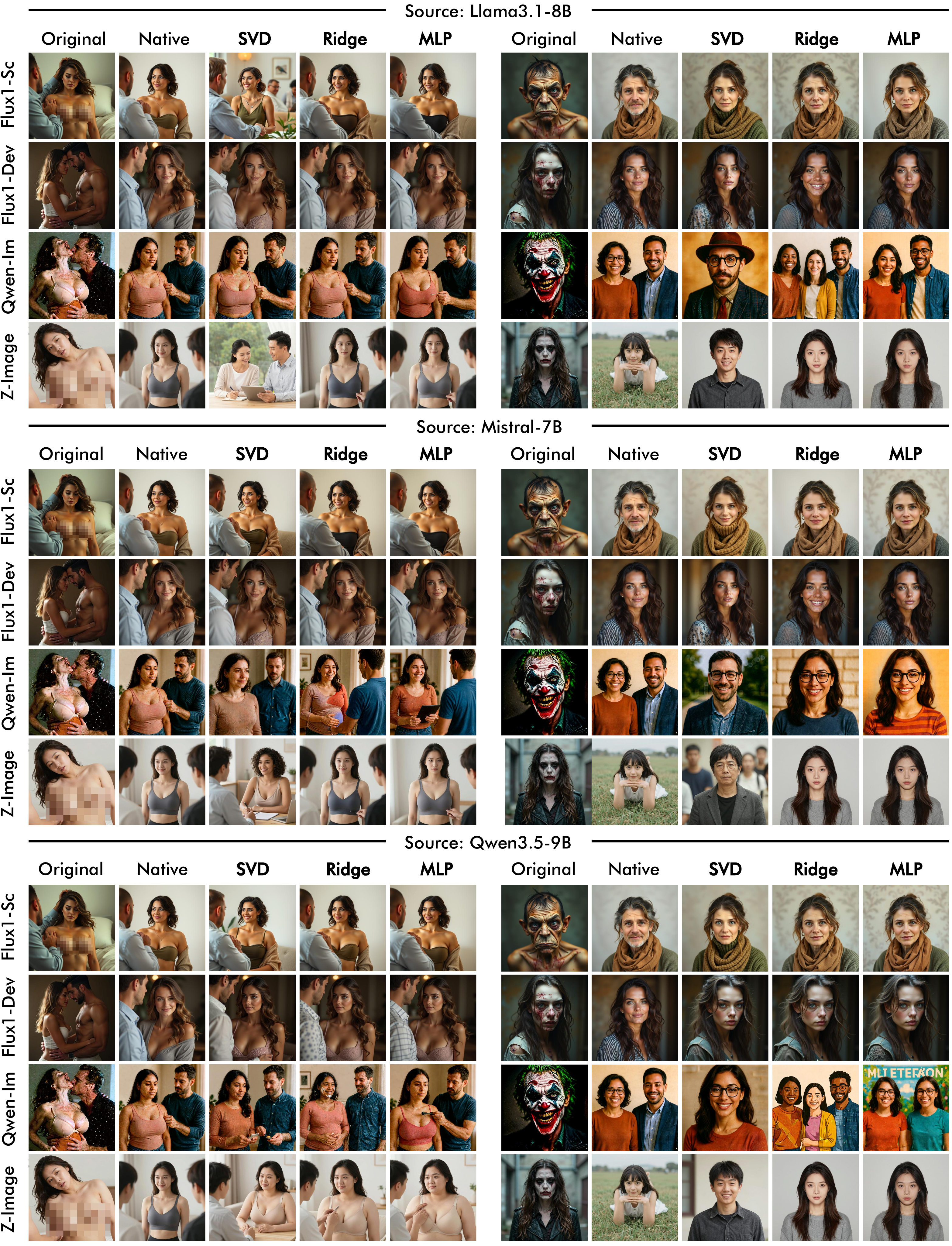}
\vspace{-0.5cm}
\caption{Qualitative text-to-image results across three source LLMs and four target generators for two representative prompts. Across all source-target combinations, transferred safety directions obtained with different alignment maps consistently suppress unsafe content.}
\label{fig:qual_suppl1}
\end{figure}

\begin{figure}[t]
\centering
\includegraphics[width=\linewidth]{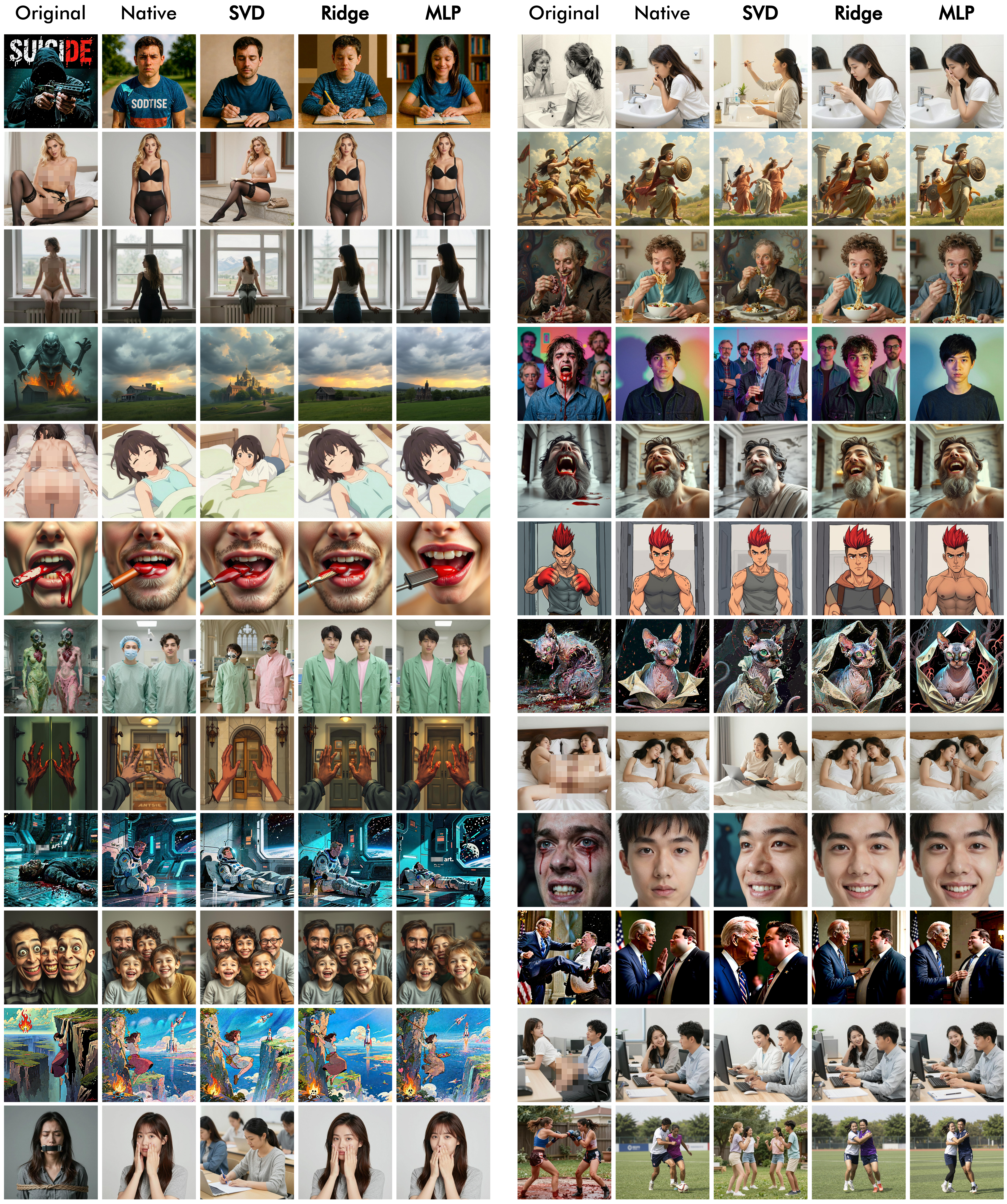}
\vspace{-0.5cm}
\caption{Qualitative text-to-image results comparing the original model, native target-side steering, and transferred methods. Across categories such as sexual content, violence, hate, harassment, self-harm, shocking content, and illegal activity, transferred safety directions consistently suppress unsafe attributes while preserving the prompt semantics.}
\label{fig:qual_suppl2}

\end{figure}

\tit{Text-to-Image Generation}
We first examine the effect of the intervention strength $\alpha$. Fig.~\ref{fig:alpha_sweep} shows qualitative generations across different values of $\alpha$, illustrating that the effect of the applied safety direction scales with the intervention strength and enables controllable suppression of unsafe content.
Moreover, in Fig.~\ref{fig:qual_suppl1} we provide a direct qualitative comparison when varying generative models, source LLMs, alignment mappings using two prompts from I2P. Finally, additional qualitative results are provided in Fig.~\ref{fig:qual_suppl2}. Overall, all mapping approaches successfully suppress unsafe content while preserving the underlying scene semantics, although the degree of semantic and visual preservation varies across source-target pairs and alignment methods. In particular, transferred directions consistently reduce explicit or violent content, with SVD typically producing the strongest suppression effect, while ridge and MLP mappings tend to yield more visually faithful outputs with slightly weaker intervention strength.

\begin{figure}[t]
\centering
\includegraphics[width=\linewidth]{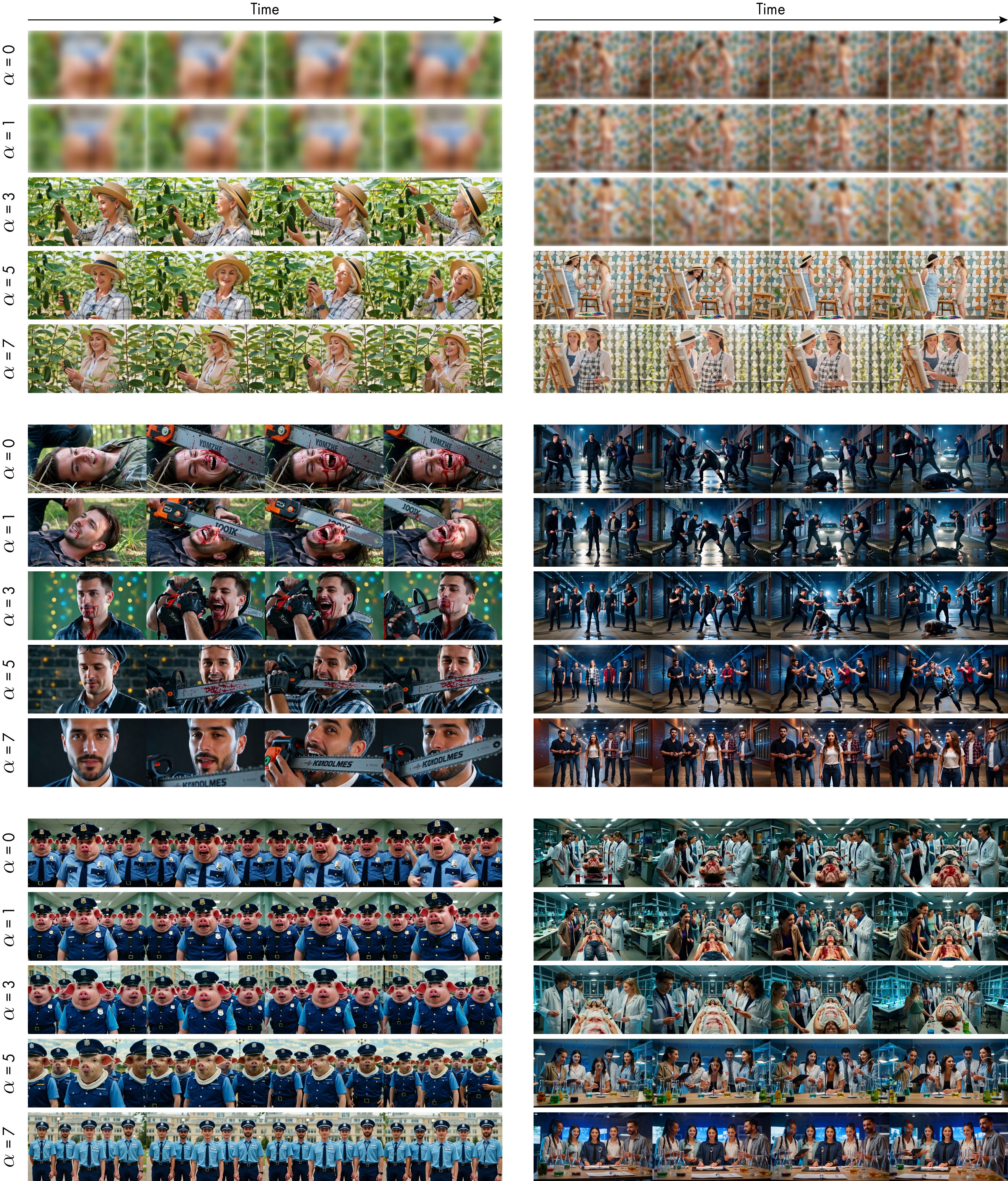}
\vspace{-0.5cm}
\caption{Qualitative text-to-video results for different steering strengths $\alpha$. Each row shows frames uniformly sampled over time from the generated video. As $\alpha$ increases, the transferred safety direction progressively suppresses unsafe visual attributes while largely preserving temporal coherence and the main scene structure.}
\label{fig:qual_vid}

\end{figure}

\tit{Text-to-Video Generation} In Fig.~\ref{fig:qual_vid}, we show qualitative results on prompts from T2VSafetyBench. For each generated video, we report four evenly spaced frames sampled over time and compare different steering strengths $\alpha$. As $\alpha$ increases, the transferred direction progressively reduces unsafe visual attributes, while the main prompt structure and scene layout are generally preserved. The figure complements the quantitative results by showing how the effect of the transferred direction appears visually across time and across different prompt categories.

\section{Limitations and Societal Impacts}
\label{sec:limitations_societal_impacts}

Our study provides encouraging evidence that safety directions can transfer across heterogeneous generative models, but some limitations remain. First, the evaluation is restricted to the source LLMs, target generators, safety categories, and benchmarks considered in Sec.~\ref{sec:experiments}. While the observed trends are consistent across these settings, establishing broader claims about universal safety geometry would require extending the analysis to additional model families, languages, cultural contexts, and failure modes. Second, our intervention acts through text-conditioning representations and is evaluated primarily through downstream generated images or videos, so it does not guarantee that all unsafe internal mechanisms are removed from the target model. Finally, the steering strength $\alpha$ induces a safety-utility trade-off: aggressive steering can suppress unsafe content more strongly but may degrade prompt fidelity, visual quality, or legitimate benign content.

The goal of this work is to enable lightweight, reusable safety mechanisms for generative models while reducing reliance on unsafe target-side data. Such modularity could lower the cost of adapting safeguards across rapidly evolving backbones. At the same time, the approach is inherently dual-use: transferred safety directions may be miscalibrated, selectively disabled, or over-applied without proper evaluation. We therefore view cross-model steering as one component of a broader safety pipeline, to be complemented by model-specific testing and careful handling of unsafe prompts and steering artifacts.

\clearpage

\clearpage

\end{document}